\documentclass[journal]{IEEEtran}
\usepackage{amsfonts}
\usepackage{amsmath}
\usepackage{graphicx}
\usepackage{float}
\usepackage{xcolor}
\usepackage{amssymb}
\usepackage{booktabs}
\usepackage{color}
\usepackage{capt-of}
\usepackage{array}
\usepackage{makecell}
\usepackage{booktabs} %
\usepackage{multirow}
\usepackage{hyperref}
\usepackage{url}
\usepackage{arydshln}
\hypersetup{
	colorlinks=true,
	linkcolor=blue,
	filecolor=gray,      
	urlcolor=red,
	citecolor=blue,
}

\begin{document}
\title{RGBT Tracking via Multi-Adapter Network \\with Hierarchical Divergence Loss}
%
%
%

\author{Andong Lu,~
		Chenglong Li,~
		Yuqing Yan,
        Jin Tang,
        and~Bin Luo
\thanks{The authors are with Anhui University.}
}

\markboth{IEEE Transactions on Image Processing}%
{Shell \MakeLowercase{\textit{et al.}}: Bare Demo of IEEEtran.cls for IEEE Journals}

\maketitle

\begin{abstract}

RGBT tracking has attracted increasing attention since RGB and thermal infrared data have strong complementary advantages, which could make trackers all-day and all-weather work.
%
%
Existing works usually focus on extracting modality-shared or modality-specific information, but the potentials of these two cues are not well explored and exploited in RGBT tracking.
In this paper, we propose a novel multi-adapter network to jointly perform modality-shared, modality-specific and instance-aware target representation learning for RGBT tracking. 
To this end, we design three kinds of adapters within an end-to-end deep learning framework. 
In specific, we use the modified VGG-M as the generality adapter to extract the modality-shared
target representations.
To extract the modality-specific features while reducing the computational complexity, we design a modality adapter, which adds a small block to the generality adapter in each layer and each modality in a parallel manner.
Such a design could learn multilevel modality-specific representations with a modest number of parameters as the vast majority of parameters are shared with the generality adapter.
We also design instance adapter to capture the appearance properties and temporal variations of a certain target. 
Moreover, to enhance the shared and specific features, we employ the loss of multiple kernel maximum mean discrepancy to measure the distribution divergence of different modal features and integrate it into each layer for more robust representation learning.
Extensive experiments on two RGBT tracking benchmark datasets demonstrate the outstanding performance of the proposed tracker against the state-of-the-art methods.

\end{abstract}

\IEEEpeerreviewmaketitle

\section{Introduction}

\IEEEPARstart{R}{GBT} tracking is an emerging topic in the computer vision community~\cite{Li2016Learning,zhu2019dense,li2019manet,zhang2019mfdimp}.
Its goal is to employ the complementary advantages of visible and thermal information to achieve robust visual tracking.
In recent years, many efforts have been devoted to promoting the progress of RGBT tracking, but there is still much research room due to the underutilization of RGB and thermal information.

Previous CNN-based works on RGBT tracking can be generally categorized into two aspects according to how they model multi-modal information.
One is to use a two-stream network to extract modality-specific features and then combine all of them using some strategies to achieve object tracking~\cite{zhang2018learning,li2018fusing,zhu2019dense,zhang2019mfdimp,2020MaCNet}.
Although the lenses of RGB and thermal modalities are with different imaging bands, their images have much correlated information such as object boundaries, spatial layout and some fine-grained textures.
Some methods~\cite{zhang2018learning,li2018fusing} do not take into consideration the collaboration of different modalities in feature learning, which might limit tracking performance.
Other methods~\cite{zhu2019dense,zhang2019mfdimp,2020MaCNet} introduce cross-modal interaction layers to capture collaboration information of different modalities, but might ignore shared information across modalities.
Therefore, many redundant parameters would be introduced.
The other one is to use a single-stream network to extract modality-shared information, i.e., use the same network parameters to extract features of all modalities~\cite{2020FANet}.
This kind of methods could model the collaborative cues of different modalities effectively, but ignore the heterogeneous properties of RGB and thermal data.
The useful single-modal information is sometimes suppressed and the tracking performance is thus degraded.
To handle these problems, we propose a novel Multi-Adapter Network (MANet) to jointly perform modality-shared, modality-specific and instance-aware feature learning in an end-to-end trained deep framework for RGBT Tracking.
Specifically, we adopt the first three convolutional layers of the modified VGG-M~\cite{vgg15iclr} as the generality adapter to extract modality-shared representations.
It should be noted that other networks like Inception Network~\cite{InceptionNet15icml} and Residual neural Network (ResNet)~\cite{ResNet} could be also applied in our framework.
We select VGG-M for its good balance between accuracy and complexity in tracking.
To improve efficiency, we introduce an adaptive RoIAlign layer~\cite{RT-MDNet18eccv} in the generality adapter to allow features of all samples to be extracted from feature maps.
One generality adapter is used to extract the features of both modalities for the modeling of the collaboration of the two modalities.

\begin{figure*}[t]
\centering  
\includegraphics[width=2\columnwidth]{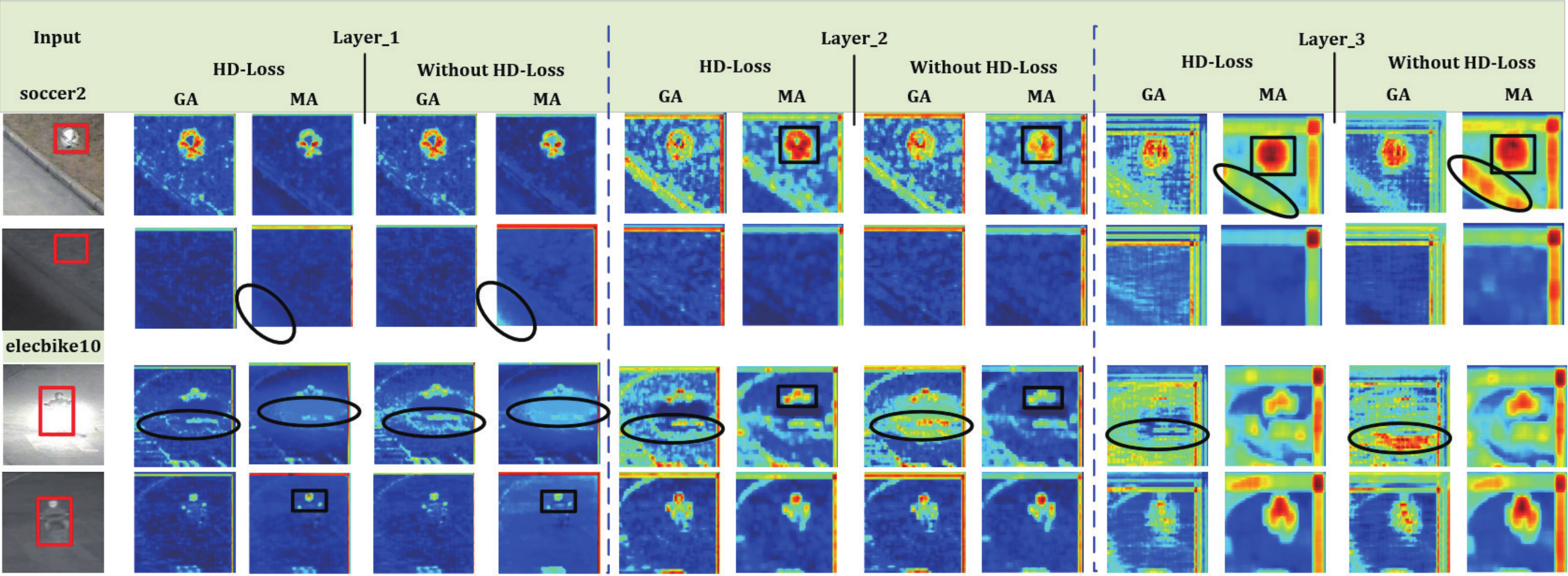}   
\caption{Illustration of the effectiveness of the hierarchical divergence loss in our modality adapter (MA) and generality adapter (GA).
We show the averaged feature maps of all layers with and without the hierarchical divergence loss, where some regions are highlighted by the black circle and black rectangle. }
\label{fig:feature_maps}
\end{figure*}

To model the heterogeneous properties of RGB and thermal sources, we design the modality adapter to extract modality-specific information.
Considering the real-time nature of visual tracking, we reduce the parameters of the modality adapter by sharing a large portion of parameters with the generality adapter. 
In specific, we add a small block which consists of a small convolution kernel (e.g., 3$\times$3 or 1$\times$1), a normalization layer and a pooling layer on the generality adapter in each layer and each modality in a parallel manner.
Although only small convolution kernels are used, our modality adapter is sufficient to encode modality-specific information as different modalities could share a large portion of their parameters and the number of the modality-specific parameters should be much smaller than the generality adapter.

To capture appearance changes and temporal variations of a certain target, we design an instance-aware adapter, which is updated online every several frames interval for the balance of accuracy and efficiency.
Our instance adapter is similar to the fully connected layers in MDNet~\cite{MDNet15cvpr}, but differ them from the following aspects.
First, we use two fully connected layers for each modality to extract its features. 
Second, we compute the modality weights to achieve quality-aware fusion of different modalities.
Finally, we concatenate the re-weighted features and then use two additional fully connected layers for target classification and regression. 
%
%

%

To improve the representation of generality and modality adapters, we want to reduce the feature differences of different modalities in generality adapter since these features should contain the shared information and increase the difference in modality adapter as they should contain modality-specific information.
Note that there are multi-layers for generality and modality adapters, and we thus design a hierarchical divergence loss (HD loss), in which each layer is embedded with a divergence loss.
To improve the robustness to outliers, we employ the multiple kernel maximum mean discrepancy~\cite{wang2016learning} to measure the distribution divergence of different modalities features.
Therefore, we minimize the divergence in generality adapter and maximize it in modality adapter in the optimization process via back propagation.

We show the effectiveness of the HD loss in Fig.~\ref{fig:feature_maps}.
The results show that HD loss is beneficial to improve the discriminative ability of GA and MA (the black rectangle), and some noises are suppressed (the black circle).
The modality-shared and modality-specific features are thus more effectively learnt using HD loss.
Note that the contrast of some feature maps decreases when using HD loss, and the reason is that
HD loss is to minimize the modality-shared feature distribution in GA and maximize the modality-specific feature distribution in MA, while the binary classification loss and instance embedding loss aim to drive discriminative feature learning of target from background in both modalities. 
Therefore, the learning of modality-shared and modality-specific features is collaboratively enhanced. 
In Fig.~\ref{fig:feature_maps}, we can see that the noise features in the black circle are suppressed in MA when using HD loss as in GA, and the target features in the black rectangle are highlighted in MA when using HD loss.

%

%
%
This paper makes the following major contributions in RGBT tracking and related applications. 
\begin{itemize}
\item It presents a novel multi-adapter framework to extract the modality-shared, modality-specific and instance-aware feature representations for robust RGBT tracking. The proposed framework is general and could be easily extended to other multimodal tasks. The source code has been released\footnote{http://chenglongli.cn/code-dataset/}.
\item It designs a parallel and hierarchical structure of the generality adapter and modality adapter and integrates the hierarchical divergence loss to establish a one-stage joint learning of modality-shared and modality-specific features.
Such a design is able to use a small number of parameters to learn powerful multilevel modality-specific representations. 
\item It designs a dynamic fusion module in the instance adapter to achieve quality-aware fusion of different source data. 
Unlike fusion strategies in existing works~\cite{2020FANet,zhu2019dense}, our fusion layer is instance-aware and thus better to capture target appearance dynamics.
\item Extensive experiments on three RGBT tracking benchmark datasets suggest that the proposed tracker achieves excellent performance against the state-of-the-art methods.
\end{itemize}

This work, called MANet++, is an extension of our previous conference version MANet~\cite{li2019manet}.
Compared with MANet, MANet++ makes the following major contributions. First, we propose a hierarchical divergence loss (HD loss) to enhance the quality of features output from modality and generality adapters. With the HD loss, we can establish a one-stage joint learning of modality-shared and modality-specific features, which avoids the risk of over-fitting in previously designed two-stage learning algorithm. 
Second, to achieve quality-aware fusion of different modalities, we design a dynamic fusion module in the instance adapter while MANet does not include any fusion scheme. 
We also make the following improvements over MANet. First, we use the RoIAlign layer to spatially align the feature map with the input image, and features of all samples are thus be extracted directly from feature map. Second, we replace the original local response normalization with the independent component to enhance the independence of neurons and eliminate redundant information in the modality adapter. 


\section{Related Work}
In recent years, more and more RGBT trackers have been proposed, and we review them from the following two aspects.

\subsection{Traditional Methods for RGBT Tracking}
Cvejic \emph{et al.}~\cite{Cvejic07cvpr} investigates the effect of pixel-level fusion of visible and infrared videos on object tracking performance.
After that, the representative works are based on sparse representation~\cite{wu2011multiple,Li2016Learning,lan2018modality,lan2019learning}, manifold ranking~\cite{Li18eccv,li2018two} and dynamic graph~\cite{li2018learning,li2019rgb}.
Early works focus on the sparse representation due to their robustness to noise and outliers. 
For example, Wu \emph{et al.}~\cite{wu2011multiple} integrate image patches from different modalities and then use a sparse representation for each sample in the target template space.
Lan \emph{et al.}~\cite{lan2019learning} propose a modality-consistency sparse representation framework and propose discriminability-consistency constrained feature template learning to learn robust feature templates for sparse representation in RGB-infrared modalities.

Following works partition the target bounding box into a set of local patches, and construct a graph to compute weights of patches.
Robust features are achieved by weighting patch features and the structured SVM is adopted for tracking.
For example, Li \emph{et al.}~\cite{Li18eccv} propose a cross-modal manifold ranking algorithm with soft consistency and noise labels to compute the patch weights.
Also, Li \emph{et al.}~\cite{li2018two} propose a two-stage modality-graphs regularized manifold ranking algorithm to mitigate the impact of inaccurate patch weights initialization.
These works, however, rely on the structure-fixed graphs, and the relations among patches are not well explored.
To handle this problem, Li \emph{et al.}~\cite{li2018learning} propose a spatially regularized graph learning to automatically explore the intrinsic relationship of global patches and local patches.
Besides, Li \emph{et al.}~\cite{li2019rgb} propose a sparse representation regularized graph learning to explore patch relations in an adaptive manner.

\subsection{Deep Learning for RGBT Tracking}
Deep learning techniques have received great success in the computer vision community, and recent works on RGBT tracking also focus on deep learning.
Li \emph{et al.}~\cite{li2018fusing} propose a two-stream convolutional neural network which uses deep neural network to learn modality-specific features, and employ correlation filter to track using the selected discriminative features.
Yang \emph{et al.}~\cite{yang2019learning} propose two local attention and global attention to train strong discriminative deep classifiers for robust RGB-T object tracking.
Zhu \emph{et al.}~\cite{2020FANet} propose a novel deep network architecture to aggregate hierarchical deep features within each modality to handle the challenge of significant appearance changes in tracking.
Zhang \emph{et al.}~\cite{xu2018relative} propose a convolutional filter containing two types, object filter and relative filters, to construct a two-layer convolutional neural network to learn sparse feature representation of RGB and thermal data for object tracking.
Zhang \emph{et al.}~\cite{zhang2019siamft} based on the fully convolutional Siamese networks propose a RGB-infrared fusion tracking method, which employs two Siamese network to extract search frame features and template frame features from each modality, and then fuse these features to generate a score map for target location. 
However, these methods employ two CNNs to extract modality-specific features, while the shared information is ignored in feature learning and some redundant parameters are also introduced.

Some works use a single network to extract both features of RGB and thermal modalities.
Zhu \emph{et al.}~\cite{zhu2019dense} propose a deep fusion method to recursively aggregate multilevel and multi-modal features, and then use the pruning algorithm to remove redundant features.
Zhang \emph{et al.}~\cite{zhang2020object} propose an attention-based deep network to adaptively fuse multilevel and multi-modal features.
However, these methods do not model modality-specific information in feature learning and the tracking performance might be limited as RGB and thermal data are usually heterogeneous.
In this paper, we take both modality-shared and modality-specific modeling into account in feature learning as well as the instance-aware fused features for robust RGBT tracking.


\begin{figure*}[t]
\centering
\includegraphics[width=2.1\columnwidth]{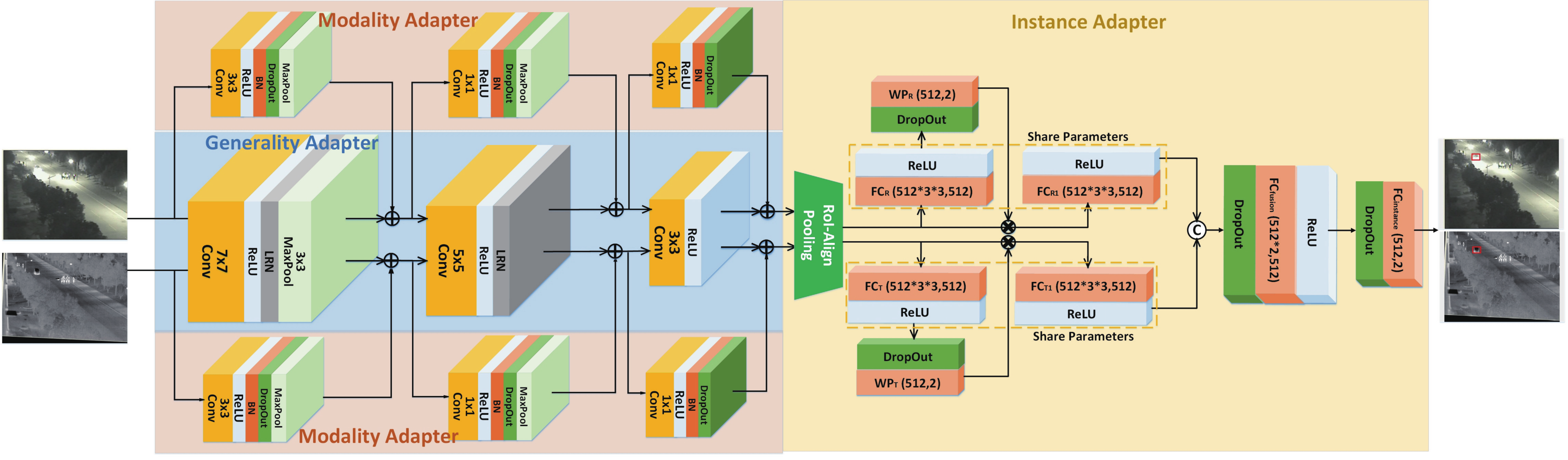}
\caption{
Overall network architecture of MANet++.
It consists of three modules: MA, GA and IA.
Herein, +, $\times$ and c denote the element-wise addition, multiplication and concatenation respectively.
$\mathit{ReLU}$, $\mathit{LRN}$ and $\mathit{BN}$ refer to the rectified linear unit, the local response normalization unit and batch normalization respectively.
In IA, $FC_R$, $FC_T$, $FC_{R1}$ and $FC_{T1}$ are fully connected layers, where $FC_R$ and $FC_{R1}$ share parameters and $FC_T$ and $FC_{T1}$ share parameters.
$WP_R$ and $WP_T$ are single fully-connected layers with 2 unit outputs, and $FC_{instance}$ is composed of $K$ such fully-connected layers.}

\label{fig::pipeline}
\end{figure*}

\section{Multi-Adapter Convolutional Network}

In this section, we will present the proposed multi-adapter network called MANet++, including network architecture, loss functions and training details.

\subsection{Network Architecture}
The pipeline of MANet++ is shown in Fig.~\ref{fig::pipeline}, in which the detailed parameter settings are presented. 
Our MANet++ consists of three kinds of network blocks, i.e., generality adapter, modality adapter and instance adapter.
The network input is two whole images from RGB and thermal modalities. 
We extract two types of features of each modality through the shared generality adapter and the modality adapter. 
Then we combine these two types of features of each modality using the element-wise addition.
Through introducing the RoIAlign layer, features of all candidate samples are directly extracted from the combined feature maps. 
Next, for each candidate, we send its features of all modalities into the instance adapter for information fusion.
Finally, we use the binary classification layer to predict the score of each candidate, and then select the candidate with the highest score as tracking result in the current frame.  

{\flushleft {\bf Generality adapter (GA)}}.
Visible spectrum and thermal infrared data are captured from cameras of different imaging bands, and thus reflect different properties of target objects.
In spite of it, they share some common information like object boundaries, spatial layout and some fine-grained textures, and thus how to model them plays a critical role in learning collaborative representations of different modalities.
However, existing works~\cite{li2018fusing,2020FANet,zhang2019mfdimp,yang2019learning} usually model different modalities separately, and thus ignore modality-shared information. 
Furthermore, separate processing of each modality would introduce a lot of redundant parameters, as different modalities should have a large portion of shared parameters.
To handle these problems, we design a generality adapter (GA) to extract shared object representations across different modalities.
There are many potential networks~\cite{vgg15iclr,ResNet} to be used for our GA, and we select the VGG-M network~\cite{vgg15iclr} for its good balance between effectiveness and efficiency.

In specific, our GA consists of the first three layers of the VGG-M network, where the convolution kernel sizes are $7\times 7\times 96$, $5\times 5\times 256$, $3\times 3\times 512$ respectively. 
The first and second layers of GA are composed of a convolutional layer, an activation function of rectified linear unit ($\mathit{ReLU}$) and a local response normalization ($\mathit{LRN}$). 
The details are shown in Fig.~\ref{fig::pipeline}.
We use the dilated convolution~\cite{chen2017deeplab} in the third layer with a dilation ratio of 3 to increase the resolution of feature maps. 
Followed by the third layer, an adaptive RoIAlign layer is employed to align feature maps spatially and produces $7\times 7$ feature maps for each sample, and then uses the max pooling layer to pool feature maps into $3\times 3$~\cite{RT-MDNet18eccv}.

{\flushleft {\bf Modality adapter (MA)}}.
As discussed above, RGB and thermal modalities are heterogeneous with different properties, and thus only using GA is insufficient for RGBT feature presentations.
To model the characteristics of each modality and make best use of the complementary advantages of RGB and thermal modalities, we need to design a sub-network to learn modality-specific feature representations. 
Recent works~\cite{li2018fusing,2020FANet,zhang2019mfdimp} use two-stream Convolutional Neural Networks (CNNs) to extract RGB and thermal features respectively.
They ignore modality-shared feature learning and usually contain abundant parameters, which might degrade tracking accuracy and efficiency respectively. 
To improve RGBT feature representations and reduce computational complexity, we propose the modality adapter (MA) that is built on GA to effectively extract modality-specific feature representations with a little computational burden.

In specific, we design a parallel network structure that includes a small convolutional kernel (e.g., 3$\times$3 or 1$\times$1) at each convolutional layer of GA.
Although only small convolutional kernels are used, our MA is able to encode modality-specific information effectively.
Since different modalities should share a large portion of their parameters, the number of modality-specific parameters should be much smaller than GA.
In particular, we develop an adaptive scheme to determine the size of the convolution kernel of MA according to the kernel size of GA. 
The kernel sizes of our MA are set to 3$\times$3 (7$\times$7 in GA), 1$\times$1 (5$\times$5) and 1$\times$1 (3$\times$3) respectively. 
The number of channels in each layer of MA and GA is consistent so that shared and the specific features can be directly added. 
Such design makes MA has only 20\% parameters of GA, which greatly reduces redundant parameters compared to two-stream networks. 
To capture more effective modality-specific information and improve generalization capability, we assign an Independent Component (IC) layer~\cite{Chen2019IClayer} in each layer of MA after the convolutional layer and $\mathit{ReLU}$ activation function, and followed by the IC layer is the max pooling layer.

Next, we explain why we can design such a parallel architecture as follows.
The feature transfer between two layers in a modality can be formulated as :
\begin{equation}
\begin{aligned}
&{\mathcal{F}_m^l}  \mathcal{=}  {\mathcal{F}_m^{{l-1}}}\ast{\mathcal{W}}\\
\end{aligned}
\label{eq::1}
\end{equation}
where ${\mathcal{F}_m^l}$ refers to the ${l}$-layer feature maps in the modality ${m}$, and ${m}$ indicates the index of one modality.
To extract the modality-shared and modality-specific features, we aim to decompose the complete parameter ${\mathcal{W}}$ into two parts, one for the modality-shared parameter ${\mathcal{W}^{GA}}$ and the other for the modality-specific parameters ${\mathcal{W}_m^{MA}}$.
To this end, we introduce a function ${\bf{diag_{\emph{S}}(\cdot)}}$ that reshapes the matrix to a new size $\emph{S*S}$ by embedding the original matrix into the center position of the new matrix and other positions are filled with 0~\cite{oneshot2016nips}. The formula is:
\begin{equation}
\begin{aligned}
&{ {\bf diag}_{\emph{S}}({\mathcal{W}_m^{MA}})}_{{wh}}=
\left\{
            \begin{array}{lcl}
             {\mathcal{W}_m^{MA}}_{ij},w=\frac{\emph{S}-a}{2}+i,h=\frac{\emph{S}-b}{2}+j. \\
			 \qquad\qquad{s.t. }{0<i<a,0<j<b.} \\             
             \\
            \qquad{ 0},\qquad {\emph{otherwise.} }
            \end{array}              
        \right.  
\end{aligned}
\label{eq::3}
\end{equation}
where $wh$ indicates the coordinates of the elements in the new matrix, and $ij$ indicates the coordinates of the elements in the original matrix.
Therefore, we can decompose ${\mathcal{W}}$ in~\eqref{eq::1} into as follows:
\begin{equation}
\begin{aligned}
&{\mathcal{W}}  \mathcal{=}  {\mathcal{W}^{GA}} + {{\bf diag}_{\emph{S}}({\mathcal{W}_m^{MA}})}
\end{aligned}
\label{eq::2}
\end{equation}
Finally, \eqref{eq::1} is equivalently expressed as follows:
\begin{equation}
\begin{aligned}
&{\mathcal{F}_m^l}  \mathcal{=}  {\mathcal{F}_m^{{l-1}}\ast{{W}^{GA}}} + {\mathcal{F}_m^{{l-1}}\ast{{W}_m^{MA}}}\\
\end{aligned}
\label{eq::4}
\end{equation}

{\flushleft {\bf Instance adapter (IA)}}.
Instance objects involve different class labels, movement patterns and appearance changes, and tracking algorithms might thus suffer from instance-specific challenges.
Furthermore, appearance of instance objects vary much over time.
Therefore, we design an instance adapter to adapt appearance changes and instance-specific challenges.
Existing methods~\cite{li2019manet,zhu2019dense} directly inherit the idea of multi-domain learning in MDNet~\cite{MDNet15cvpr}.
Different from MDNet, our instance adapter (IA) first uses two fully connected layers for each modality, and then predicts modality weights to achieve quality-aware fusion of different modalities.
There are two major reasons why we choose the fusion position in the first fully connected layer. 
First, the parameters of IA are updated online to capture appearance dynamics of target, and thus we integrate two modalities in IA to achieve instance-aware fusion. 
Second, integrating two modalities in other layers would introduce more parameters, which affect computational speed and also easily lead to overfitting. 
We also verify this choice in experiments.

In specific, IA is composed of eight fully connected ($\mathit{FC}$) layers, named as $\mathit{FC_R, FC_{R1},}$ $\mathit{FC_T, FC_{T1}, WP_R, WP_T, FC_{fusion}}$ and $\mathit{FC_{instance}}$ with the output sizes of 512, 512, 512, 512, 2, 2, 512 and 2 respectively. 
Herein, to reduce parameters, $\mathit{FC_R, FC_{R1}}$ and $\mathit{FC_T, FC_{T1}}$ share common parameters, as shown in Fig.~\ref{fig::pipeline}.
Except for $\mathit{FC_R}$ and $\mathit{FC_T}$, other fully connected layers include a $\mathit{Dropout}$ operation.
Besides $\mathit{WP_R, WP_T}$ and $\mathit{FC_{instance}}$ layers also employ $\mathit{Softmax}$ to calculate the positive and negative scores of samples, and other fully connected layers include an activation function $\mathit{ReLU}$.
$\mathit{FC_R}$ and $\mathit{FC_T}$ are used to extract features of RGB and thermal sources separately, and $\mathit{WP_R}$ and $\mathit{WP_T}$ are employed to predict the positive score $\mathcal{P}_m$ and negative score $\mathcal{N}_m$ ($m=1,2,...,M$) respectively.
The modality weights are computed by the following equation:
\begin{equation}
\begin{aligned}
& \mathcal{\eta}_m = \Omega ( \dfrac{1}{n}|\sum_{i=0}^n (\mathcal{P}_m^i - \mathcal{N}_m^i)| )\quad n=0, 1,2 \ldots ,255
\end{aligned}
\label{eq::5}
\end{equation}
where $\mathcal{P}_m^i$ and $\mathcal{N}_m^i$ represent the positive and negative scores of the $i$-th sample in the $m$-th modality. 
$\Omega$ is the $\mathit{Sigmoid}$ function, which is used to normalize the modality weights $\mathcal{\eta}_m$ to a range of 0 to 1.
We use the modality weights to re-weight features output from RoIAlign layer, and then re-encode these feature maps by $\mathit{FC_R (FC_T)}$ layer.
Finally, the re-encoded features of RGB and thermal modalities are concatenated, and the $\mathit{FC_{fusion}}$ layer is used to fuse modal features.  
The final $\mathit{FC_{instance}}$ is to build a new FC layer for each instance target, which is used to achieve the adaptation of the instance target, similar to MDNet~\cite{MDNet15cvpr}.
In the training phase, $\mathit{FC_{instance}}$ will build an equal number of branches based on the number of sequences trained for multi-domain learning.
During online tracking, $\mathit{FC_{instance}}$ will be removed and replaced with a binary classification layer with softmax cross-entropy loss, and rebuilt once in each sequence.
Therefore, we use the newly created $\mathit{FC_{instance}}$ layer to initially model the target in the current sequence, and update IA to adapt to changes of the target over time to achieve robust tracking.
\subsection{Loss Function}
Our network includes three kinds of loss functions including hierarchical divergence loss, binary classification loss and instance embedding loss.
The hierarchical divergence loss is based on the multiple kernel maximum mean discrepancy (MK-MMD), and we thus first review it for the sake of clarity.

{\flushleft {\bf Review: MK-MMD}}.
As pointed out in~\cite{gretton2012optimal, wang2016learning}, given a set of independent observations from two distributions $p$ and $q$, the two-sample test accepts or rejects the null hypothesis $H_0: p = q$,  which measures the distance between the samples based on the values of the test.
In the topological space $\mathcal{X}$ with a reproducing kernel $\mathit{k}$, we define a reproducing kernel Hilbert space as $\mathcal{T}_k$.
The mean embedding of distribution $\mathit{p}$ in the reproducing kernel Hilbert space $\mathcal{T}_k$ is a unique element $\mathit{\mu_{k}(p)}$~\cite{berlinet2011reproducing}:  
\begin{equation}
\begin{aligned}
& \mathbf{E}_{x\sim p}\mathit{f(x)} = {\langle \mathit{f},\mu_{k}(p)\rangle}_{\mathcal{T}_k},
\qquad \forall \mathit{f} \in \mathcal{T}_k
\end{aligned}
\label{eq::6}
\end{equation}
Based on the Riesz representation theorem, when the kernel function $\mathit{k}$ is Borel-measurable and $\mathbf{E}_{x\sim p}\mathit{k}^{1/2}(x,x) < \infty$, the mean embedding $\mathit{\mu_{k}(p)}$ exists.
%

%
In fact, we calculate the $\mathcal{T}_k$-distance between the mean embedding $p$ and $q$ as the maximum mean discrepancy (MMD) between the Borel probability measures $\mathit{p}$ and $\mathit{q}$.
An expression for the squared MMD is as follows:
\begin{equation}
\begin{aligned}
& \varphi \mathit{(p,q)} = \Vert \mathit{\mu_{k}(p)} - \mathit{\mu_{k}(q)} \Vert_{\mathcal{T}_k}^2
\end{aligned}
\label{eq::7}
\end{equation}
Since MMD is strongly correlated with its kernel function $\mathit{k}$, there may be contradictory results for different kernel functions.
To handle this problem, Gretton \emph{et al.}~\cite{gretton2012optimal} propose a multiple kernel maximum mean discrepancy (MK-MMD) in a two-sample test, which selects the kernel function to maximize the testing power, and minimize the Type II error (false acceptance $\mathit{p=q}$) with a given upper boundary of type I error (false rejection $\mathit{p=q}$). 
In domain adaptation~\cite{long2015learning,qin2019pointdan}, they employ MK-MMD to improve test performance by generating kernel functions that belong to the kernel family.
Therefore, the multiple kernel function $\mathit{k}$ is a linear combination of a set of positive definite functions $\mathit{ \{ k_u \}_{u=1}^d }$, i.e.
\begin{equation}
\begin{aligned}
&\mathcal{K:}= \{  \mathit{k} = \mathit{\sum_{u=1}^d }{\beta_u}{k_u},\sum_{u=1}^d{\beta_u}=D; \forall u \in \{1,\ldots,d \}    \}
\end{aligned}
\label{eq::8}
\end{equation}
where $\mathit{D} > 0$, $\beta_u \geqslant 0;$ and each $k \in \mathcal{K}$ is uniquely in $\mathcal{T}_k$, based on the assumption that the kernel is bounded, $|k_u|\leq K$, $\forall u \in \{ 1,\ldots, d \}$.
%

{\flushleft {\bf Hierarchical divergence loss}}.
Due to the different imaging principles between different modal images, it is difficult to directly measure their similarity using Euclidean distance which is very sensitive to outliers.
However, when the distributions of the two modalities are determined to be similar, outliers with significantly different appearance can be tolerated at the same time.
Thus, we can solve this problem by treating different modalities as two samples obeying different distributions.
We pursue to make modality-shared features in two modalities with similar distributions and modality-specific features with different distributions.
There are many information theory techniques that can be used to calculate the similarity between distributions, such as KL divergence, entropy, and mutual information.
However, these existing methods tend to use bias-correction strategies, sophisticated space-partitioning, and density estimation, which are difficult to apply to high-dimensional data.
Hence, we choose MK-MMD to evaluate the similarity in this work.
Therefore, from~\eqref{eq::7} and~\eqref{eq::8}, we can use MK-MMD method to measure the distance between two distributions and formulate it as follows:
\begin{equation}
\begin{aligned}
&\mathit{\psi(p,q)}=\Vert \mathit{\mu_{k}(p)} - \mathit{\mu_{k}(q)} \Vert_{\mathcal{T}_k}^2 = \sum_{u=1}^{d}{\beta_u}{\psi_u(p,q)}
\end{aligned}
\label{eq::9}
\end{equation}
where $\mathit{\psi_u(p,q)}$ is the $\mathbf{MMD}$ for the kernel function $\mathit{k_u}$.

In specific, we output the features of each layer of GA and MA, and then calculate the modality-shared features and modality-specific features in each level separately by the following formula:
\begin{equation}
\begin{aligned}
&\mathit{\psi^j(GA_{rgb},GA_t)}= \dfrac{2}{b}\sum_{i=1}^{b/2}{H_k(u_i)},  \qquad 1<i<b
\\
&H_k(u_i)= k(GA_{rgb}^{2i-1}, GA_{rgb}^{2i}) + k(GA_{t}^{2i-1}, GA_{t}^{2i}) \\
&- k(GA_{rgb}^{2i-1}, GA_{t}^{2i})- k(GA_{t}^{2i-1}, GA_{rgb}^{2i}),
\end{aligned}
\label{eq::10}
\end{equation}
where $\mathit{b}$ is the batch size, the $\mathit{GA_{rgb}^i}$ and $\mathit{GA_{t}^i}$ indicate RGB and thermal feature maps output from GA respectively.
$\mathit{\psi^j(GA_{rgb},GA_t)}$ is denoted as the unbiased estimating of MK-MMD between the modality-shared features of the $\mathit{j}$-th layer.
%
Also, similar to~\eqref{eq::10}, we can obtain unbiased estimates between the output features of MA, written as $\mathit{\psi^j(MA_{rgb},MA_t)}$.
During the training phase, we want to minimize $\mathit{\psi^j(GA_{rgb},GA_t)}$ and maximize $\mathit{\psi^j(MA_{rgb},MA_t)}$. 
This is because in our framework, the distribution similarity between modality-shared features is expected to be as large as possible, while the distribution similarity between modality-specific features is expected to be as small as possible. 
Thus, the loss function we designed is shown below:
\begin{equation}
\begin{aligned}
&\mathit{L_{hd}}=  \sum_{j=1}^3{\mathit{\psi^j(GA_{rgb},GA_t)}} -  \sum_{j=1}^3{\mathit{\psi^j(MA_{rgb},MA_t)}}
\end{aligned}
\label{eq::11}
\end{equation}
As a result of this loss-driven, we can learn modality-specific features and modality-shared features through one-step training.
In addition, supervised training enables our model to fully mine the characteristics of each modality and improve the generalization.
{\flushleft {\bf Binary classification loss}}.
In the framework of tracking by detection~\cite{MDNet15cvpr,RT-MDNet18eccv}, the most important loss function is binary classification loss. 
The key to detection-based strategies is to distinguish between foreground and background categories.
Following MDNet, to learn the representations that distinguish target from background, we employ the loss function of binary cross entropy (BCE) to drive the discriminative learning of target and background.
%
%

In our network, we calculate the scores of the single modality and the fused one separately.
From the outputs of $\mathit{WP_R}$, $\mathit{WP_T}$ and $\mathit{FC_{instance}}$, 2D binary classification scores are recorded as $\mathit{S_R}$ , $\mathit{S_T}$ and $\mathit{S_{fusion}}$, which are formulated as follows:
\begin{equation}
\begin{aligned}
&\mathit{L_{fusion}}=  -\dfrac{1}{n} \sum_{i=1}^n\sum_{c=1}^2 {[y_i^c]_{d}} \cdot \log(\sigma([S_{fusion}^c]_{d}))\\
&\mathit{L_{R}}=  -\dfrac{1}{n} \sum_{i=1}^n\sum_{c=1}^2 {y_i^c} \cdot \log(\sigma(S_{R}^c))\\
&\mathit{L_{T}}=  -\dfrac{1}{n} \sum_{i=1}^n\sum_{c=1}^2 {y_i^c} \cdot \log(\sigma(S_{T}^c))\\
\end{aligned}
\label{eq::12}
\end{equation}
where ${y_i}\in\{0,1 \}$ is a one-hot vector of the ground-truth label, $c$ represents positive samples $(c=1)$ or negative samples $(c=0)$, and $\sigma$ is the $\mathit{Softmax}$ operation. $\mathit{[\cdot]_d}$ represents the output of the $\mathit{d}$-th $\mathit{FC_{instance}}$ layer.

Therefore, the final binary classification loss for our network is formulated as:
\begin{equation}
\begin{aligned}
&\mathit{L_{cls}}=  \mathit{L_{fusion}} + \lambda_1\mathit{L_{R}} + \lambda_2\mathit{L_{T}}
\end{aligned}
\label{eq::13}
\end{equation}
where $\lambda_1$ and $\lambda_2$ are the balance factors, in this paper we set them as $\lambda_1$ = $\lambda_2$ = 0.5 for offline training and $\lambda_1$ = $\lambda_2$ = 1 for online learning.

{\flushleft {\bf Instance embedding loss}}.
The binary classification loss function tries to distinguish target from background in each domain, which makes it weak to distinguish between targets in different domains,
especially when an object is a target in one domain and a background in another domain.

Following RT-MDNet~\cite{RT-MDNet18eccv}, our algorithm adds a constraint, which embeds target from different videos to be apart from each other.
We can implement this constraint with the loss function as follows:
\begin{equation}
\begin{aligned}
&\mathit{L_{inst}}=  -\dfrac{1}{n} \sum_{i=1}^n\sum_{d=1}^D {[y_i^+]_{d}} \cdot \log(\sigma([S_{fusion}^+]_{d}))
\end{aligned}
\label{eq::14}
\end{equation}
where $D$ is the number of domains (i.e video sequence) in a training dataset.
Note that this loss function only works on positive samples denoted by $+$ in~\eqref{eq::14}.
Under the influence of this loss function, the target score is enhanced in the current domain and suppressed in other domains.
Moreover, based on this loss-driven model, similar targets can be distinguished more effectively during testing.

{\flushleft {\bf Overall loss}}.
The loss function of our network is designed as follows:
\begin{equation}
\begin{aligned}
&L_{off} = L_{cls} +  \nu_1 L_{inst} + \nu_2 L_{hd}
\end{aligned}
\label{eq::15}
\end{equation}
where $\nu_1$ and $\nu_2$ are the hyper-parameters that control the importance of loss terms, and herein we set $\nu_1$ = 0.1.

\subsection{One-stage Learning Algorithm}
It should be noted that existing datasets not only contain ground-truths of each modality, they also provide high-aligned common ground-truths for both modalities. Therefore, in our work, we employ these shared ground-truths for training.
In the offline training phase, the whole network is trained in a one-stage end-to-end manner.
We use the stochastic gradient descent (SGD) algorithm~\cite{bottou2012stochastic} to train our model.
The specific details of training are set as follows.
We construct a mini-batch with the samples collected from a training sequence for each iteration. 
This mini-batch includes 256 positive and 768 negative examples in 8 frames randomly selected  from a single sequence, e.g., sampling 32 positive and 96 negative samples in each frame.
Herein, the criterion for selecting positive samples is that the Intersection over Union (IoU) of the sampling boxes and the ground-truth boxs is greater than 0.7,  and the IOU of the negative samples is less than 0.5.
We initialize our GA sub-network with the parameters of VGG-M~\cite{vgg15iclr} and randomly initialize the parameters of other sub-networks, and then use the RGBT dataset to fine-tune them for tracking.
Note that when testing on GTOT~\cite{Li2016Learning}, we fine-tune network using RGBT234~\cite{li2018fusing}, and vice versa.
We train the network until the model converges, where the learning rate of all layers is 0.0001, the weight decay is fixed to 0.0005. 
To more flexibly control the mining of modality-shared features and modality-specific features, we set different values at different stages of training:
$\nu_2$ = 1 in the first 200 iterations, $\nu_2$ = 0.1 between 200 and 500 rounds, and $\nu_2$ = 0.01 after 500 rounds.
In this experiment, the specific settings for the $L_{hd}$~\eqref{eq::11} term are as follows.
We use $d$ different Gaussian kernels (i.e. $k_u(p,q)=e^{-\Vert p-q \Vert ^{2/{\sigma_u}}}$) to build the kernel function in
~\eqref{eq::10} by a linear combination.
For improving the efficiency and performance, we set $d$=11, which is enough to drive modality-shared and modality-specific learning, and set $\sigma_u$ to $2^{u-6} (u=1,\ldots ,11)$.
Considering that different kernels should have different weights, and thus the setting of the parameter $\beta_u$ in~\eqref{eq::9} is the same as in~\cite{gretton2012optimal}.
In the online training phase, including initial training and online update.
To model the target instance, we recreate a new $F_{instance}$ branch for each test sequence, and fine-tune the IA of the first frame of each tracking video by using the first frame information. 
Specifically, we collect 500 positive samples and 5000 negative samples, which meet the IoUs standard developed during the offline training phase.
Initial training uses the samples collected from the first frame to train our network by 50 iterations with a learning rate of 0.001 for $FC_{instance}$ and others are 0.0001 in IA.
Moreover, we apply the bounding box regression technique~\cite{MDNet15cvpr} to improve the prediction smoothness at the target scale, and train the regressor only in the first frame.
In the subsequent frames, we draw positive samples with IoU greater than 0.7 and negative samples with IoU less than 0.3 at the estimated target location.
We save these sample feature representations from outputs of the adaptive RoIAlign layer to avoid redundant computations in the online tracking phase.
These samples are used as the training dataset for online update to maintain the robustness and adaptability of the model.
Online update consists of two complementary update mechanisms, namely long-term update and short-term update.
Short-term updates will be performed after the target score of the current frame is below the threshold, here we set the threshold to zero, while long-term updates are executed every 10 frames~\cite{MDNet15cvpr}. 
It is worth noting that we only use the loss function~\eqref{eq::13} in online training to achieve the online adaptation of our model.
%


\section{Online Tracking}

During the tracking process, we fix all parameters of GA and MA. 
We replace the last fully connected layer in instance adapter with a new one to fit the target instance of each RGBT video sequence.
Our model receives the first pair of RGBT frames with ground truth bounding box, and then performs initial training as described above.
In the subsequent frames, long-term and short-term updates are  performed according to the rules described above.
For obtaining a pair of RGBT input frames at time $\mathit{t}$, we take Gaussian sampling centered on the previous tracking result ${\mathit X}_{\mathit{t}-1}$ at time $\mathit{t}$-1, and collect 256 candidate regions as ${\mathit{ x^i_t}}$.
We use these candidate regions as input to our IA.
Next, we first obtain their classification scores based on single-modality sample features, and then calculate the weight of each modality according to~\eqref{eq::5}.
Then, we use the weight assigned to each candidate region to re-encode the $\mathit{FC_R}(FC_T)$  and use the concatenation operation to fuse modality features.
Finally, we employ the $\mathit{FC_{fusion}}$ layer to encode the fused features and use the $\mathit{FC_{instance}}$ to obtain the classification score of each sample.
Herein, the positive and negative scores of each sample are denoted as $f^+(x^i_t)$ and $f^-(x^i_t)$, respectively. 
We select the candidate region sample with the highest score as the tracking result ${\mathit {X_t}}$ at time $\mathit{t}$, and the formula expression is as follows:
\begin{equation}
{\mathit{X}}_t^{\ast}=\mathop{ \arg\max}_{i=0,...,255}\ \mathrm{\emph{f}^+({\mathit{x}^i_t)}}
\label{eq::16} 
\end{equation}
It is worth noting that when $f^+(x^i_t) > 0.5$, we will use the bounding box regression model to adjust the position and scale of the target.
When the  $f^+(x^i_t) < 0$, the short-term update will start.
Long-term updates are performed with 10 frames interval.
%

\section{Performance Evaluation}
In this section, we will compare our MANet++ with current popular tracking algorithms, including RGB trackers and RGBT trackers.
We also verify the effectiveness of the major components in the proposed algorithm. 

\subsection{Evaluation Data and Metrics}
In this paper, we evaluate our MANet++ on three large-scale benchmark datasets.

{\flushleft \bf GTOT dataset}.
GTOT dataset~\cite{Li2016Learning} contains 50 spatially and temporally aligned pairs of RGB and thermal infrared video sequences under different scenes and conditions.
The dataset is labeled with a bounding box for each frame, and 7 additional challenge attributes are labeled to evaluate different RGBT tracking algorithms for attribute-based analysis.
We employ the widely used tracking evaluation metrics, including precision rate (PR) and success rate (SR) for quantitative performance evaluation.
In specific, PR is the percentage of frames whose output location is within the threshold distance of the ground truth value, and we compute the representative PR score by setting the threshold to be 5 and 20 pixels for GTOT and RGBT234 datasets respectively (since the target objects in GTOT are generally small). 
SR is the percentage of the frames whose overlap ratio between the output bounding box and the ground truth bounding box is larger than the threshold, and we calculate the representative SR score by the area under the curve.

{\flushleft \bf RGBT234 dataset}.
RGBT234 dataset~\cite{li2019rgb} consists of 234 spatially and temporally aligned RGBT video sequences.
The longest video sequence contains about 4,000 frames, and the entire dataset has a total of 200,000 frames.
Moreover, this dataset has rich challenging factors such as motion blur, camera moving, illumination, deformation and occlusion.
These challenges are labeled separately for a more comprehensive evaluation of different RGBT tracking algorithms.
As RGBT234 dataset contains ground-truths of each modality, following existing works~\cite{li2019rgb}, we employ the maximum PR (MPR) and maximum SR (MSR) metrics for fair evaluation.
Specifically, for each frame, we compute the Euclidean distance mentioned in PR on both RGB and thermal modalities, and adopt the smaller distance to compute the precision. we also set the threshold to be 20 pixels in RGBT234 and 5 pixels in GTOT to obtain the representative MPR.
Similar to MPR, we define maximum success rate (MSR) to measure tracking results. By varying the threshold, the MSR plot can be obtained, and we employ the area under curve of MSR plot to define the representative MSR.

{\flushleft \bf VOT-RGBT2019 dataset}.
VOT-RGBT2019 dataset~\cite{vot-rgbt2019} contains 60 RGBT video sequences selected from RGBT234 dataset~\cite{li2019rgb}, with a total of 20,083 frames. 
Different from the above metrics, we follow the VOT protocol to evaluate different tracking algorithms. 
Note that in VOT protocol, when evaluated algorithms lose the target, the corresponding ground-truth will be used to re-initialize algorithms.
Three evaluation metrics, Expected Average Overlap (EAO), robustness (R) and accuracy (A), are used. 
\subsection{Evaluation on GTOT Dataset}

\begin{figure}[t]
	\centering
	\includegraphics[width=1\columnwidth,height=3.65cm]{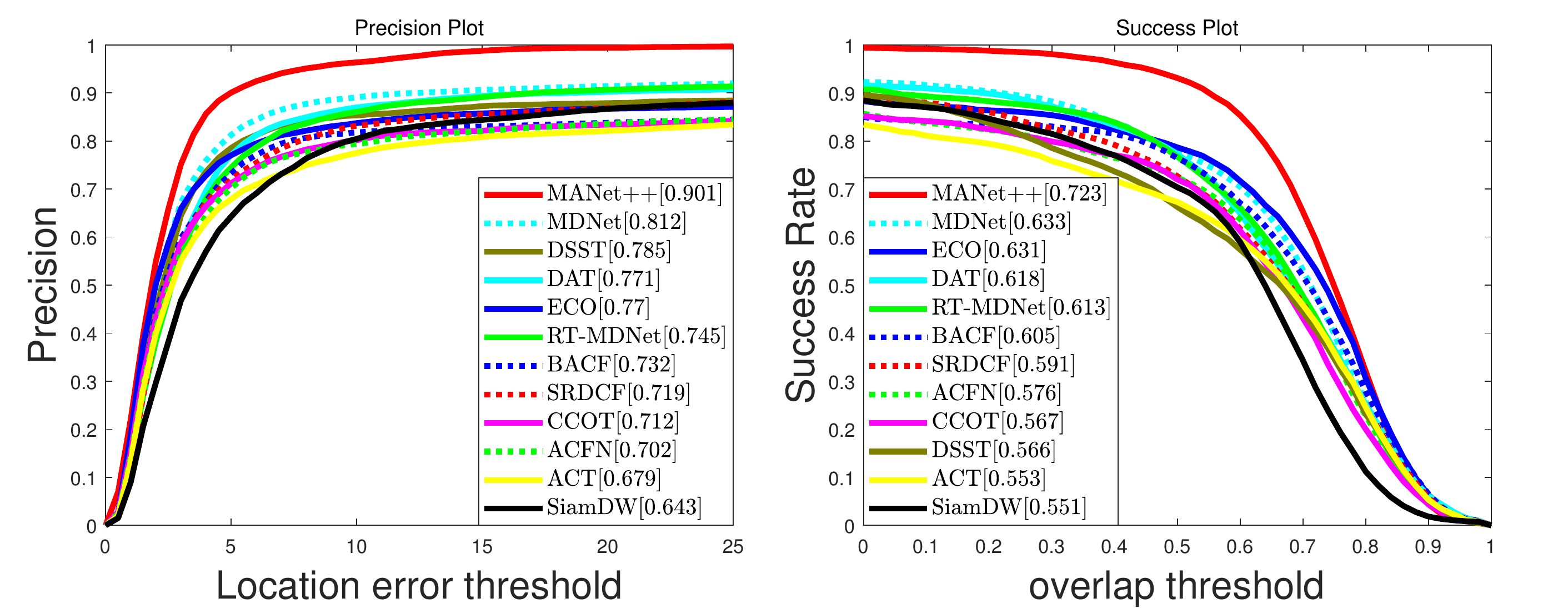}
	\caption{Comparison between our algorithm with RGB trackers on GTOT dataset, where the representative PR and SR scores are presented in the legend.}
	\label{fig:GTOT_RGB_results}
\end{figure}

On the GTOT dataset, we first compare with 11 RGB trackers, including 
ECO~\cite{ECO17cvpr}, DAT~\cite{Pu2018Deep}, RT-MDNet~\cite{Jung2018Real}, C-COT~\cite{C-COT16eccv},  ACT~\cite{Chen2018Real} and SiamDW~\cite{Zhipeng2019Deeper}, SRDCF~\cite{martin2015SRDCF}, BACF~\cite{kiani2017BACF}, ACFN~\cite{choi2017ACFN}, DSST~\cite{martin2014DSST}, MDNet~\cite{MDNet15cvpr}.
The results are shown in Fig.~\ref{fig:GTOT_RGB_results}.
Our tracker outperforms MDNet~\cite{MDNet15cvpr}, DSST~\cite{martin2014DSST} and DAT~\cite{Pu2018Deep} with $8.9\%/9.0\%$, $11.6\%/15.7\%$ and  $13.0\%/10.5\%$ in PR/SR, respectively.
From the results, we can see that our approach significantly outperforms all RGB trackers on GTOT dataset~\cite{Li2016Learning}.
It fully demonstrates that our method is able to make best use of thermal modalities to boost tracking performance.

\begin{figure}[t]
	\centering
	\includegraphics[width=1\columnwidth,height=3.65cm]{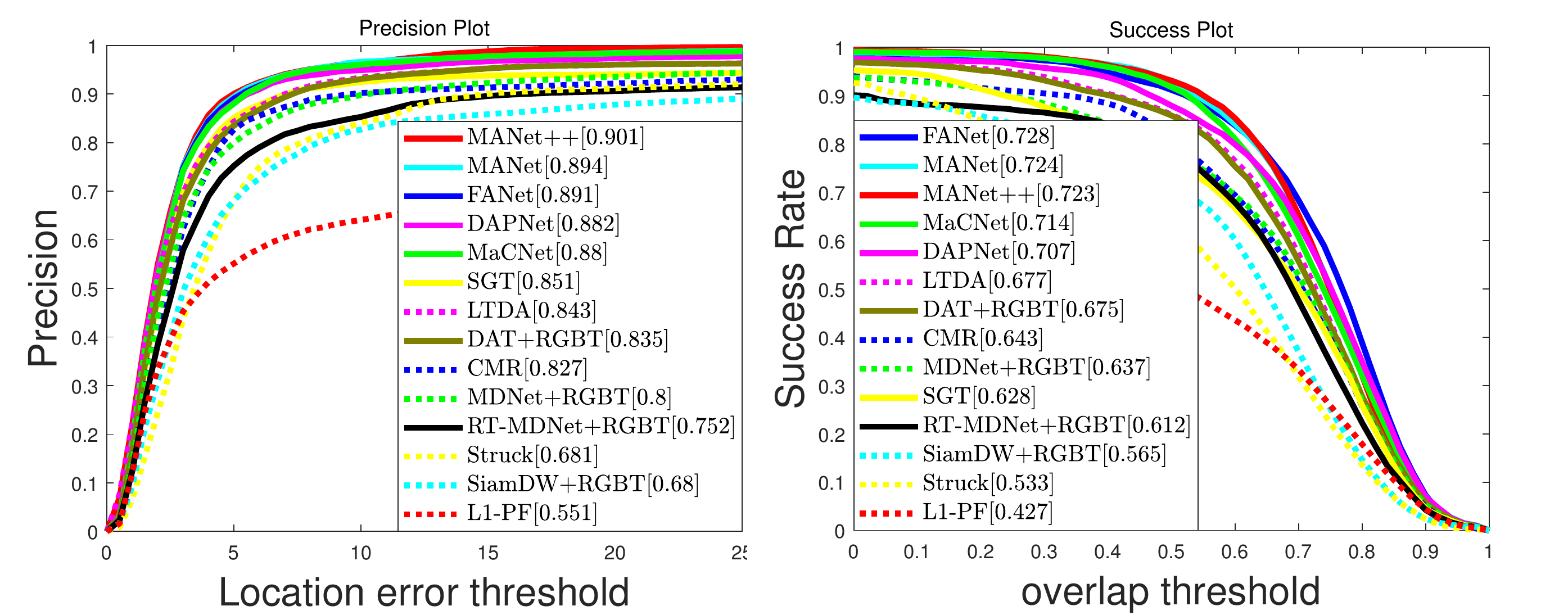}
	\caption{Comparison between our algorithm and RGBT trackers on GTOT dataset, where the representative PR and SR scores are presented in the legend.}
	\label{fig:GTOT_RGBT_results}
\end{figure}

We also compare our approach with 13 state-of-the-art RGBT trackers, some of which are from the GTOT benchmark.
Since there are not many existing deep-based tracking methods in the RGBT tracking field, we extend some RGB algorithms to RGBT ones.
Specifically, the extended methods are to add the thermal modality data as an additional channel of the RGB modality and input it into RGB trackers.
Here, DAT~\cite{Pu2018Deep}+RGBT, MDNet~\cite{MDNet15cvpr}+RGBT, RT-MDNet~\cite{RT-MDNet18eccv}+RGBT, Struck~\cite{Stuck11iccv}+RGBT, SiamDW~\cite{Zhipeng2019Deeper}+RGBT are extended algorithms.
Other RGBT trackers include MANet~\cite{li2019manet}, MaCNet~\cite{2020MaCNet}, FANet~\cite{2020FANet}, DAPNet~\cite{zhu2019dense}, SGT~\cite{Li17rgbt210}, LTDA~\cite{yang2019learning}, L1-PF~\cite{wu2011multiple} and CMR~\cite{Li18eccv}
From Fig.~\ref{fig:GTOT_RGBT_results} we can see that our tracker exceeds most of RGBT algorithms.

Our tracker outperforms MANet~\cite{li2019manet}, FANet~\cite{2020FANet} and DAPNet~\cite{zhu2019dense} with $0.7\%$, $1.0\%$ and  $1.9\%$ in PR, respectively.
However, our method has slightly low SR compared with MANet~\cite{li2019manet} and FANet~\cite{2020FANet}.
%
%
Although MANet++ is slightly worse than MANet in SR on GTOT dataset (0.723 \emph{vs.} 0.724), it is better in PR (0.901 \emph{vs.} 0.894). 
Moreover, our MANet++ is 8 times faster than MANet in speed on GTOT dataset.
These results demonstrate the effectiveness of the added hierarchical divergence loss and RoIAlign layer.
%
%
%
Note that FANet~\cite{2020FANet} adopts features of all layers for target classification and regression, in which shadow features are very important for accurate target localization. 
%
While we only use the highest layer features but achieve superior performance on PR metric over FANet on GTOT dataset, which fully demonstrates the effectiveness of our tracker.
%

%

%
%

\subsection{Evaluation on RGBT234 Dataset}
To further evaluate the effectiveness of our method, we perform a series of experiments on a large-scale dataset RGBT234~\cite{li2019rgb}, including overall performance, challenge-based performance and visual results.

\begin{figure}[t]
\centering
\includegraphics[width=1\columnwidth]{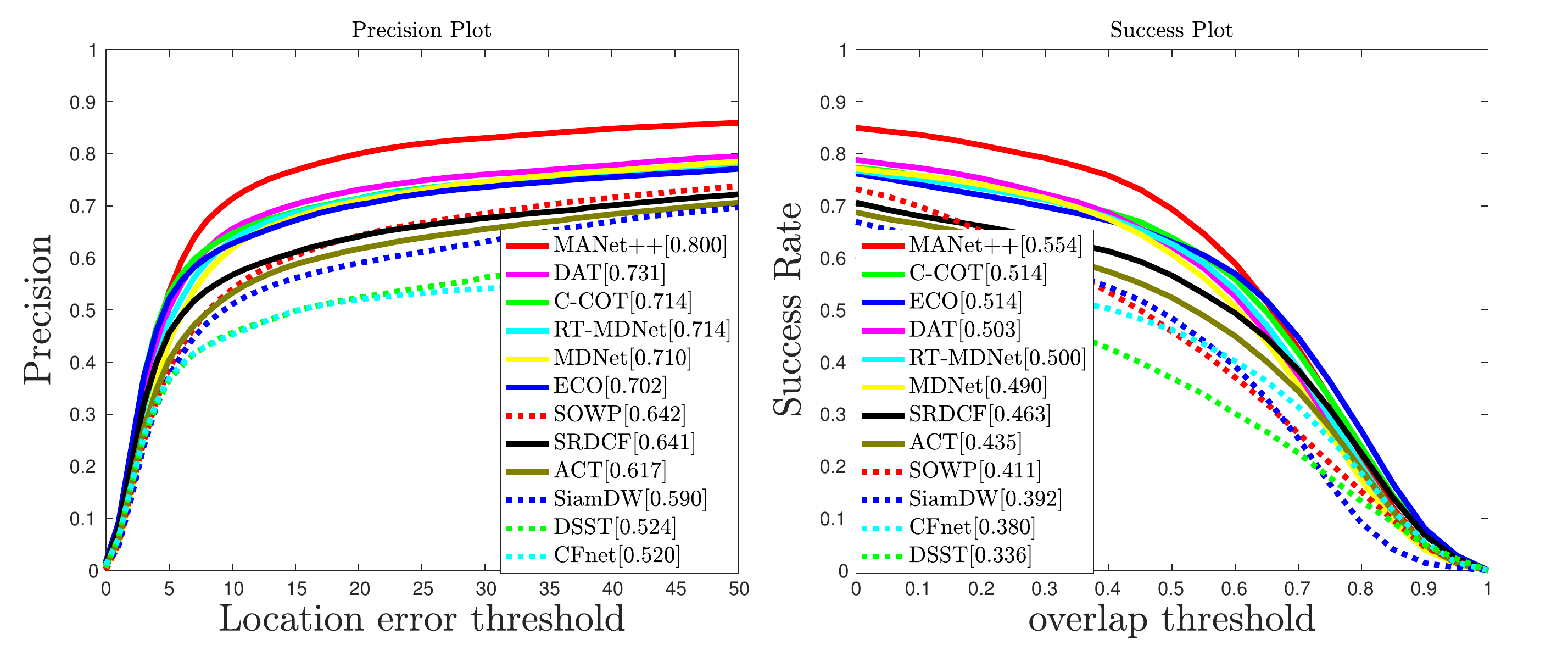}
\centering
\caption{Comparison of our algorithm and RGB trackers on RGBT234 dataset, where the representative PR and SR scores are presented in the legend.}
\label{fig:RGBT234_RGB_results}
\end{figure}

\begin{figure}[t]
\centering
\includegraphics[width=1\columnwidth]{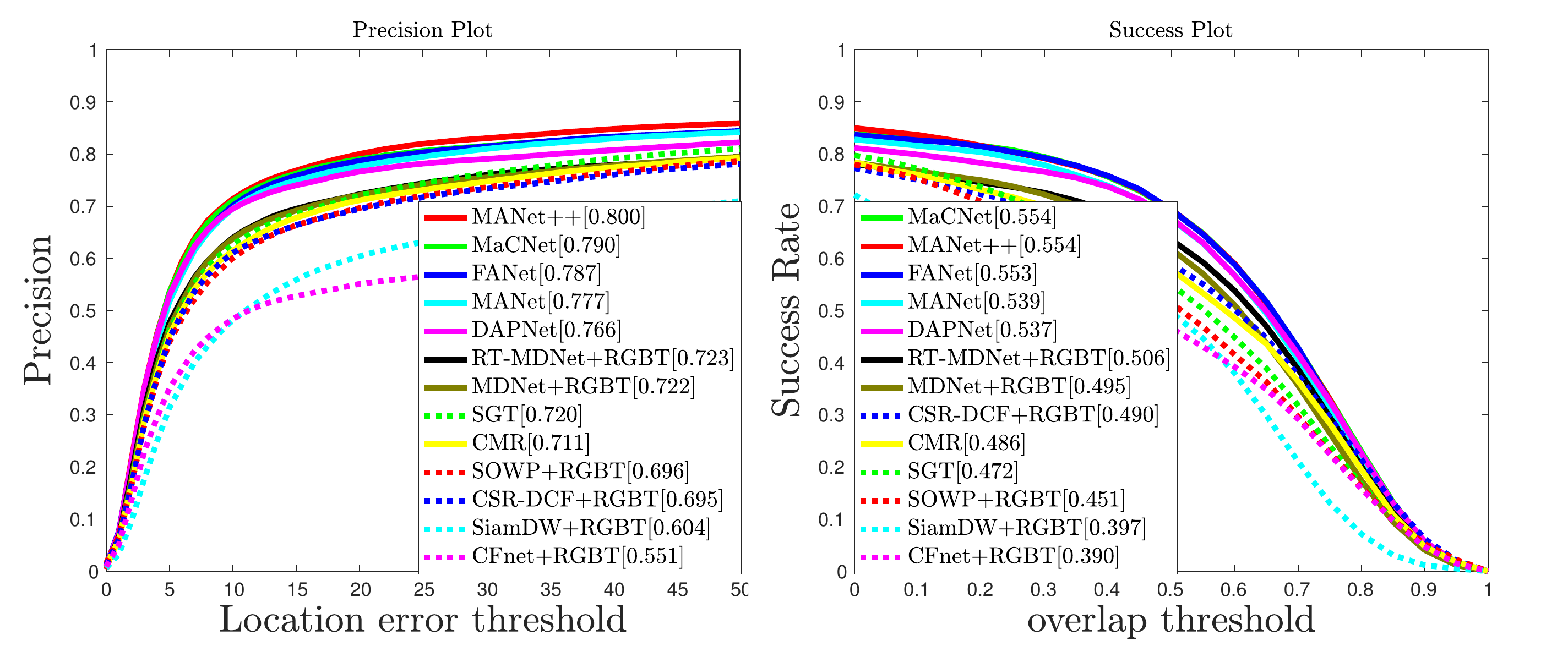}
\centering
\caption{Comparison of our algorithm and the RGBT trackers on RGBT234 dataset, where the representative PR and SR scores are presented in the legend.}
\label{fig:RGBT234_RGBT_results}
\end{figure}

{\flushleft \bf{Overall performance}}.
We compare our method with 10 state-of-the-art RGBT trackers as shown in Fig.~\ref{fig:RGBT234_RGBT_results}, including SGT~\cite{Li17rgbt210}, FANet~\cite{2020FANet},MacNet~\cite{2020MaCNet}, DAPNet~\cite{zhu2019dense}, MANet~\cite{li2019manet}, MDNet~\cite{MDNet15cvpr}+RGBT, CSR-DCF~\cite{dcf-csr16cvpr}+RGBT, SiamDW~\cite{Zhipeng2019Deeper}+RGBT, RT-MDNet~\cite{Jung2018Real}+RGBT, CMR~\cite{Li18eccv}, CFNet~\cite{valmadre2017CFnet}+RGBT and SOWP~\cite{kim2015sowp}+RGBT.
We also compare with the current advanced 11 RGB trackers as shown in Fig.~\ref{fig:RGBT234_RGB_results}. They are ECO~\cite{ECO17cvpr}, DAT~\cite{Pu2018Deep}, RT-MDNet~\cite{Jung2018Real}, C-COT~\cite{C-COT16eccv},  ACT~\cite{Chen2018Real}, CSR-DCF~\cite{dcf-csr16cvpr}, SOWP~\cite{kim2015sowp}, DSST~\cite{martin2014DSST}, CFnet~\cite{valmadre2017CFnet} and SiamDW~\cite{Zhipeng2019Deeper}.
%
	From the results we can see that our MANet++ outperforms all other trackers on RGBT234 dataset~\cite{li2018fusing} in all metrics.
%
It fully demonstrates the effectiveness of our algorithm and the importance of thermal modality information.
In particular, our MANet++ has the highest performance, i.e., $80.0\%$ and $55.4\%$ in PR and SR respectively.
It not only achieves $6.9\%$ performance gains in PR over the second best RGB tracker DAT, and $4.0\%$ performance gains in SR over the second best RGB tracker C-COT, but also outperforms the baseline RGBT tracker MANet over $2.3\%/1.5\%$ in PR/SR.

{\flushleft \bf{Challenge-based performance}}.
RGBT234 dataset includes 12 challenge attribute labels, including no occlusion (NO), partial occlusion (PO), heavy occlusion (HO), low illumination (LI), low resolution (LR), thermal crossover (TC), deformation (DEF), fast motion (FM), scale variation (SV), motion blur (MB), camera moving (CM) and background clutter (BC).
The evaluation results are shown in Table~\ref{tb::AttributeResults}.
From the results we can see that our proposed method achieves best in most challenges, including the challenges of PO, HO, LI, LR, TC, DEF, FM, SV, CM and BC.
In the evaluation of the LR challenge, our algorithm has a $6\%$ performance improvement over the second place in the PR.
It shows that our model can make full use of the information of the two modalities.
Furthermore, for the challenges of PO, LI, TC, DEF, CM and BC, tracking performance is improved by about $3\%$.
It also demonstrates that our algorithm has strong discriminative ability of target features.
Compared with MANet, our MANet++ has a remarkable improvement in the PO, LI and TC challenges.
%


\begin{table*}[t]
	\centering
	\linespread{1.3}
	\setlength{\belowcaptionskip}{0.1cm}
	\caption{Attribute-based PR/SR scores (\%) on RGBT234 dataset compared with eight RGBT trackers. The best and second results are in $\color{red} red$ and $\color{blue} blue$ colors, respectively.}
\smallskip 
	\begin{tabular}{ c | c  c  c  c  c  c  c  c | c c }
	   \hline
		\textbackslash\ &  \thead[c]{SOWP+RGBT} &CFNet+RGBT &CMR & SGT &MDNet+RGBT &RT-MDNet+RGBT & DAPNet & MANet & MANet++  \\ \hline
		NO &{86.8}/53.7 & 76.4/56.3 & 89.5/61.6 & {87.7}/55.5 & 86.2/61.1 & 85.5/61.3 & $\color{red}90.0$/64.4 & 88.7/$\color{blue}64.6$ & 
		$\color{blue}89.8 $/$\color{red}65.4$  \\[2pt] 
		 
		PO & 74.7/48.4 & 59.7/41.7 & 77.7/53.5 & 77.9/51.3 & 76.1/51.8 &74.5/52.6 & ${\color{blue}82.1}$/${\color{blue}57.4}$ &  81.6/56.6 &  ${\color{red} 85.2}$/${\color{red} 59.3}$ \\[2pt] 
		
		HO & 57.0/37.9 & 41.7/29.0 & 56.3/37.7 & 59.2/39.4 & 61.9 /42.1& 64.1/43.9 & 66.0/45.7 & $\color{blue}68.9$/$\color{blue}46.5$
		& ${\color{red} 70.4}$/${\color{red} 47.1}$ \\[2pt]
		
		LI & 72.3/46.8 & 52.3/36.9 &74.2/49.8 & 70.5/46.2 & 67.0/45.5 &58.9/39.8 & $\color{blue}77.5$/$\color{blue}53.0$ &  76.9/51.3&  $\color{red}81.1$/${\color{red} 55.1}$ \\[2pt]
		
		LR & 72.5/46.2 & 55.1/36.5 & 68.7/42.0 & 75.1/47.6 &
		$\color{blue}75.9$/$\color{blue}51.5$ &70.8/48.7 &75.0/51.0 & 75.7/$\color{blue}51.5$	& ${\color{red} 82.3}$/${\color{red} 54.5}$ \\[2pt]
		
		TC & 70.1/44.2 & 45.7/32.7 & 67.5/44.1 &76.0/47.0 & 75.6/51.7 &76.0/${\color{blue} 55.8}$& $\color{blue}76.8$/54.3 & 75.4/54.3 & ${\color{red} 80.3}$/${\color{red} 57.6}$\\[2pt]
		
		DEF & 65.0/46.0 & 52.3/36.7 & 66.7/47.2 & 68.5/47.4 & 66.8/47.3 & 69.0/49.4 & 71.7/51.8 & $\color{blue}72.0$/$\color{blue}52.4$
		  & ${\color{red} 75.3}$/${\color{red} 53.5}$ \\[2pt]
		
		FM & 63.7/38.7 & 37.6/25.0 & 61.3/38.2 & 67.7/40.2 &58.6/36.3& 64.6/42.7& 67.0/44.3 & $\color{blue}69.4$/$\color{blue}44.9 $
	
		  &${\color{red} 70.0}$/${\color{red} 45.3}$  \\[2pt]
		
		SV &66.4/40.4 & 59.8/43.3 & 71.0/49.3 & 69.2/43.4 &73.5/50.5 & 75.1/53.4& $\color{blue}78.0$/$\color{blue}54.2$ & 77.7/$\color{blue}54.2$
	
	    & ${\color{red} 78.9}$/${\color{red} 55.4}$\\[2pt]
		
		MB & 63.9/42.1 & 35.7/27.1 & 60.0/42.7  & 64.7/43.6 & 65.4/46.3 & 65.8/47.9& 65.3/46.7 & $\color{red}72.6$/$\color{red}51.6$
	
		 &${\color{blue} 72.0}$/${\color{blue} 51.1}$\\[2pt]
		CM & 65.2/43.0 & 41.7/31.8 & 62.9/44.7  & 66.7/45.2 & 64.0/45.4 & 65.1/46.9& 66.8/47.4 & $\color{blue}71.9$/$\color{blue}50.8$

		&${\color{red}74.7}$/${\color{red} 52.3}$\\[2pt]
		
		BC & 64.7/41.9 & 46.3/30.8 & 63.1/39.7 & 65.8/41.8& 64.4/43.2& 66.4/43.5& 71.7/48.4 & $\color{blue}73.9$/$\color{blue}48.6$
		  &${\color{red} 76.7}$/${\color{red} 49.1}$\\[2pt] \hline
		
		\thead[c]{ALL} &69.6/45.1 & 55.1/39.0 &71.1/48.6 & 72.0/47.2 &72.2/49.5& 72.3/50.6& 76.6/53.7 & $\color{blue}77.7$/$\color{blue}53.9$
		 & ${\color{red} 80.0}$/${\color{red} 55.4}$ \\ \hline
	 \end{tabular}
 	
	\label{tb::AttributeResults}
\end{table*}

\begin{figure*}[t]
	\centering
	\includegraphics[width=2.1\columnwidth]{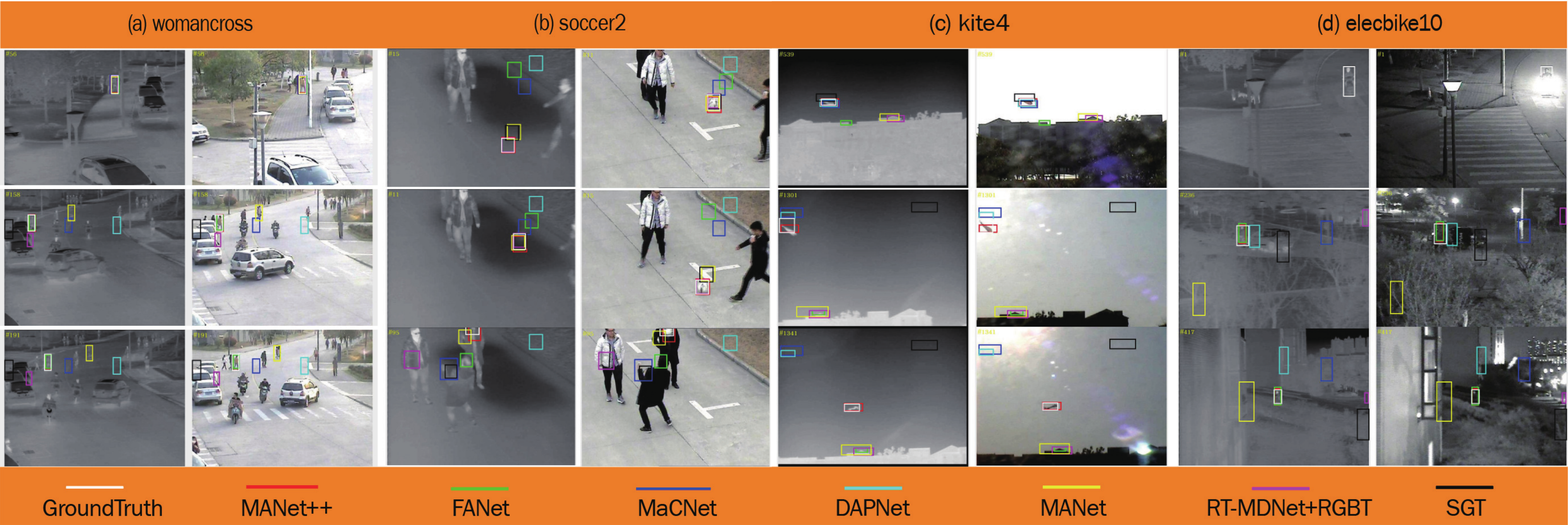}
	\caption{Qualitative comparison of MANet++ against other state-of-the-art trackers on four video sequences.
	}
	\label{fig::Visual_examples}
\end{figure*}

{\flushleft \bf{Visual comparison}}.
In Fig.~\ref{fig::Visual_examples}, we compare MANet++ with six advanced RGBT algorithms, including FANet~\cite{2020FANet}, MaCNet~\cite{2020MaCNet}, DAPNet~\cite{zhu2019dense}, MANet~\cite{li2019manet}, RT-MDNet~\cite{Jung2018Real}+RGBT and SGT~\cite{Li17rgbt210}, on four sequences.
In the sequence $womancross$, compared with other methods, MANet++ can accurately locate the target and perform better on occlusion and background cluster challenges.
From the sequence $soccer2$, our algorithm can better handle occlusion and 
thermal crossover challenges.
%
In the sequence $kite4$, it can be seen that our algorithm sufficiently suppresses the interference of high illumination.
It is worth noting that in the sequence $elecbike10$, the initial target in RGB modality contains strong illumination information, which makes most algorithms dominated by this information. Therefore, when the illumination becomes normal, most algorithms are model drift and lose the target. 
However, our method can well suppress this modality-specific noise information as shown in Fig.~\ref{fig:feature_maps}, which ensures more accurate target location.
It suggests that our algorithm can sufficiently mine the information of two modalities.
Overall, through the above comparison, our algorithm is better able to deploy the information of two modalities to deal with complex challenges in real scenarios.

\begin{figure}[t]
	\centering
	\includegraphics[width=\columnwidth]{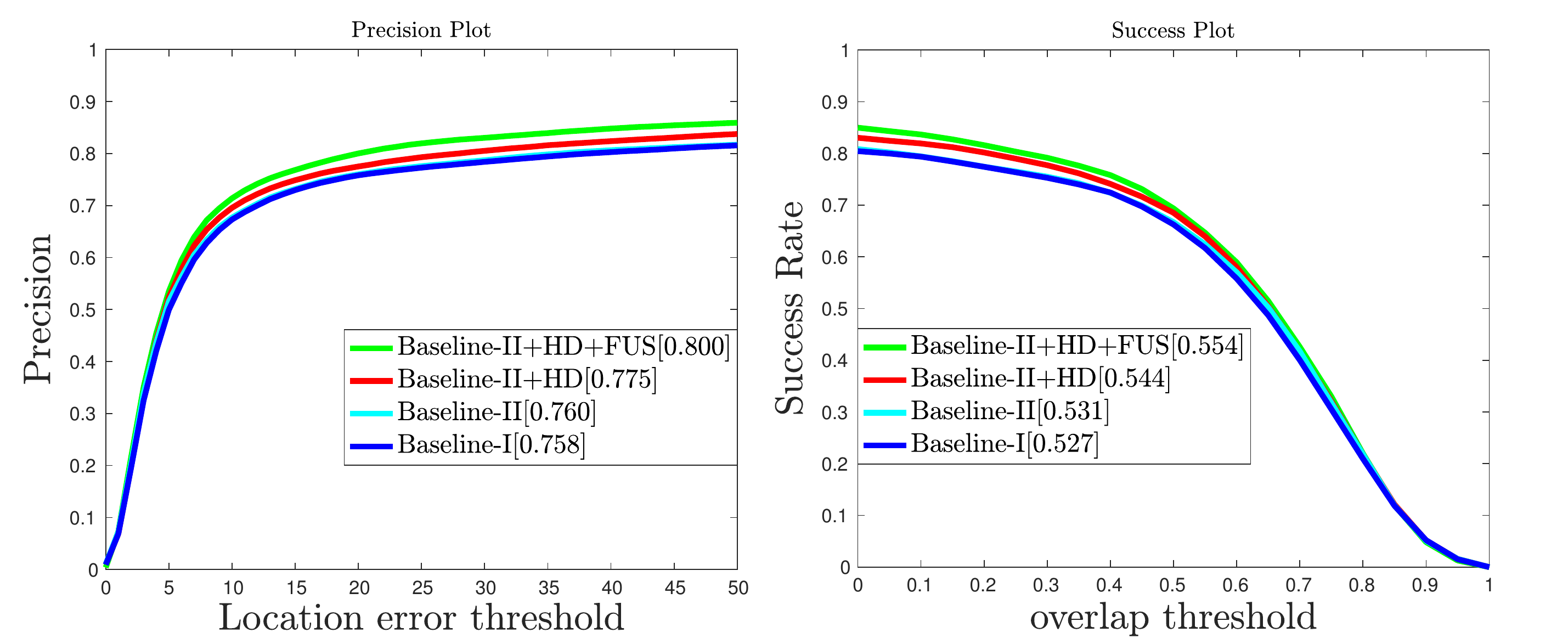}
	\centering
	\caption{Comparison results of MANet++ and its variants on RGBT234 dataset, where the representative PR and SR scores are presented in the legend.}
	\label{fig:234_component}
\end{figure}


\begin{table*}[!h]
	\caption{Comparison results on VOT-RGBT2019 dataset.}
	\label{tb::VOT_result2}
	\centering
	\footnotesize %
	\begin{tabular}{c| c c c: c c: c c c }
		\hline
		Method & MaCNet &FANet &mfDiMP &MDNet+RGBT  & MANet &RT-MDNet+RGBT & MANet++-RoIAlign  &MANet++ \\
		\hline
		Params(Mb) &56.720 &146.628 &670.719 &17.312 &27.802 &17.311 &28.163  &28.163 \\
		\hline
		FPS       	  &0.8 &19 &10.3	&3.6   &3.1	  &35.5 &3.1  &25.4 \\
		\hline
		A($\uparrow$) &0.5451 &0.4724 &0.6019 &0.5707  &0.5823 &0.4817  &0.5821  &0.5092  \\
		R($\uparrow$) &0.5914 &0.5078 &0.8036 &0.5806  &0.7010 &0.3760  &0.7259  &0.5379  \\
		\hline
		EAO  &0.3052 &0.2465 &0.3879 &0.2827 &0.3463 &0.2136 &0.3635 &0.2716 \\
		\hline
	\end{tabular}
\end{table*}

\subsection{Evaluation on VOT2019-RGBT Dataset}

To more comprehensively evaluate the effectiveness of our algorithm over other state-of-the-art methods, we present the performance comparison in Table~\ref{tb::VOT_result2}, including MANet~\cite{li2019manet}, FANet~\cite{2020FANet}, MaCNet~\cite{2020MaCNet}, mfDiMP~\cite{zhang2019mfdimp}, MDNet~\cite{MDNet15cvpr}+RGBT and RT-MDNet~\cite{RT-MDNet18eccv}+RGBT.
%
Since most of compared algorithms are based on MDNet, we implement a variant of MANet++ for evaluation, called MANet++-RoIAlign, that removes RoIAlign layer in MANet++.
From the results we can see that our MANet++-RoIAlign has comparable performance against mfDiMP~\cite{zhang2019mfdimp} and outperforms other state-of-the-art methods including MANet~\cite{li2019manet}, FANet~\cite{2020FANet} and MaCNet~\cite{2020MaCNet}. 
%
It demonstrates the effectiveness of the added HD loss and IC layer on VOT-RGBT2019 Dataset.

MANet++ significantly outperforms RT-MDNet+RGBT and FANet on VOT-RGBT2019 dataset, which demonstrates the effectiveness of our MANet++. 
However, the performance of MANet++ is inferior to MANet.
The major reason is that RoIAlign operation might lead to weak representation ability of deep features in representing low-resolution objects, as demonstrated by the comparison of MANet++-RoIAlign, MANet++ and MANet.
Note that MANet++ advances MANet in the following three aspects. 
First, it is 8 times faster in speed and achieves real-time performance. 
Second, it is better in PR (0.901 \emph{vs.} 0.894) on GTOT dataset, although slightly worse than MANet in SR (0.723 \emph{vs} 0.724). 
Finally, it improves the performance by 2.3\%/1.5\% in PR/SR on RGBT234 dataset. 

\subsection{Ablation Study}
To prove the effectiveness of the major components adopted in the proposed method, we implement four variants and perform comparative experiments on RGBT234.
The four variants are: 1) Baseline-I, that adopts two-stage learning algorithm like in MANet based on the RT-MDNet, and the normalization layer of MA adopts local response normalization ($\mathit{LRN}$);
2) Baseline-II, that replaces $\mathit{LRN}$ in Baseline-I with the Independent Component~\cite{Chen2019IClayer} layer and others are unchanged;
3) Baseline-II+HD, that integrates the hierarchical divergence loss in Baseline-II and uses one-stage learning algorithm;
4) Baseline-II+HD+FUS, that incorporates the quality-aware fusion scheme in Baseline-II+HD.
The comparison results on RGBT234 are shown in Fig.~\ref{fig:234_component}.

From the results, we can make the following conclusions:
a) Using the IC layer instead of $\mathit{LRN}$ is helpful to improve tracking performance.
b) The hierarchical divergence loss enables modality adapter and generality adapter to fully mine effective modality-shared and modality-specific features.
c) The fusion strategy in IA is beneficial to achieve quality-aware fusion of different modalities and thus improve tracking performance clearly. 

\begin{table}[!h]
	\caption{Comparison of performance of our method against several variants on RGBT234 dataset.}
	\label{tb::fusion_position_234}
	\centering
	\footnotesize %
	\begin{tabular}{c| c c }
		\hline
		Methods & PR & SR   \\
		\hline
		MANet++ &80.0  &55.4 \\
		\hline
		MANet++$_{early}$ &77.5 &54.4 \\
		MANet++$_{late}$ &76.0 &53.7\\
		\hline
	\end{tabular}
\end{table}

\subsection{Impact of Fusion Position}

	To show the influence of different fusion positions, we design several variants of the proposed method, and the results on RGBT234 dataset are shown in Table~\ref{tb::fusion_position_234}.
Herein, MANet++$_{late}$ denotes that we perform fusion at the second FC layer and MANet++$_{early}$ at the last convolution layer.
	From the results we can see that MANet++ achieves the best performance, which demonstrates the choice at the first FC layer in our MANet++. 
	%

%

In addition, the EAO of MANet significantly outperforms MDNet~\cite{MDNet15cvpr}+RGBT and MaCNet~\cite{2020MaCNet}.

\subsection{Efficiency Analysis}

We implement our algorithm on the  PyTorch 0.4.1 platform with 2.1 GHz Inter(R) Xeon(R) CPU E5-2620 and NVIDIA GeForce RTX 2080Ti GPU with 11GB memory.
We report our tracker the runtime and the size of parameters against some state-of-the-art RGBT trackers in Table~\ref{tb::VOT_result2}. 
%
%
From Table~\ref{tb::VOT_result2} we can see that MANet++ has faster tracking speed compared with MaCNet~\cite{2020MaCNet}, FANet~\cite{2020FANet}, and has comparable or higher performance on VOT-RGBT2019 dataset.
In specific, MANet++ is about 1.34 times faster than FANet and 31.75 than MaCNet, and has a much small number of parameters than FANet and MaCNet. 
%

In addition, MANet~\cite{li2019manet} has lower speed than FANet~\cite{2020FANet}.
The major reason is that MANe does not introduce the RoIAlign layer.
Moreover, we can see that the efficiency of MANet is very close to MDNet+RGBT and very faster than MaCNet, where MDNet+RGBT is to add the thermal modality data as an additional channel of the RGB modality and inputs this four-channel data into a single network MDNet. 
It suggests that MANet has comparable efficiency with the baseline MDNet+RGBT and higher efficiency than two-stream network MaCNet.
The similar observations are drawn from the size of network parameters in Table~\ref{tb::VOT_result2}.
Therefore, our MANet is able to use a small number of parameters to efficiently learn powerful multilevel modality-specific representations.
%

%
%

\begin{table}[!h]
	\caption{Comparison of performance and s peed of our method against several variants on GTOT and RGBT234 dataset.}
	\label{tb::RuntimeResults_GTOT}
	\centering
	\footnotesize %
	\begin{tabular}{c| c c c: c c c }
		\hline
		& \multicolumn{3}{c:} {GTOT} &\multicolumn{3}{:c} {RGBT234}  \\
		\hline
		Methods & PR & SR & FPS & PR & SR & FPS  \\
		\hline
		MANet++ &90.1  &72.3 &27.3 &80.0  &55.4 &25.4\\
		\hline
		MANet++-IC &87.2 &70.0 &27   &78.5  &55.3  &24.6  \\
		MANet++-RoIAlign &90.3 &73.1 &3.3	 &80.6   &55.2  &3.1	\\
		MANet &89.4 &72.4 &3.5 &77.7 &53.9 &3.1 \\
		\hline
	\end{tabular}
\end{table}

To verify the influence of several components on tracking speed and performance, we design several variants shown in Table~\ref{tb::RuntimeResults_GTOT} on GTOT and RGBT234 dataset.
Herein, MANet++-IC is the version that removes IC layers from all modality adapters in MANet++, and MANet++-RoIAlign is the version that removes RoI Align layer in MANet++.
From the results we can see that RoI Align layer plays a crucial role in tracking speed and accuracy.
In previous MANet, each candidate (256 candidates in total) needs to pass through the network to extract features, which is time consuming. 
Through introducing the ROIAlign layer, features of all candidates can be directly extracted from feature maps of input image and the tracking speed is thus improved greatly.
The IC layers have a slight impact on tracking speed and accuracy.

\section{Conclusion}
In this paper, we propose a novel multi-adapter neural network to learn powerful RGBT representation and fusion for object tracking.
The network consists of three types of adapters, including generality adapter, modality adapter and instance adapter.
In particular, the generality adapter and the modality adapter have a parallel architecture and share most of parameters for effective and efficient design.
We also introduce the hierarchical divergence loss to improve features extracted from generality and modality adapters.
Moreover, we design a quality-aware fusion scheme and embed it into the instance adapter.
Extensive experiments on two benchmark datasets demonstrate the effectiveness and efficiency of the proposed tracking method.
In future work, we will explore more modal sources such as depth and near infrared data in our framework for more robust tracking, and study deeper networks (e.g., ResNet) for more powerful representations of generality and modality adapters.

\bibliographystyle{IEEEtran}
\bibliography{mybibfiles}



%

\begin{IEEEbiography}{Michael Shell}
Biography text here.
\end{IEEEbiography}

\begin{IEEEbiography}{John Doe}
Biography text here.
\end{IEEEbiography}


\begin{IEEEbiographynophoto}{Jane Doe}
Biography text here.
\end{IEEEbiographynophoto}




\end{document}